\documentclass[conference]{IEEEtran}
\IEEEoverridecommandlockouts
\usepackage{cite}
\usepackage{amsmath,amssymb,amsfonts}
\usepackage{algorithmic}
\usepackage{graphicx}
\usepackage{textcomp}
\usepackage{xcolor}
\usepackage{times}
\usepackage{latexsym}
\usepackage{booktabs}
\usepackage{algorithm}
\usepackage{algorithmic}
\usepackage{amsmath}
\usepackage{multirow}
\usepackage{graphicx}
\usepackage{float}

\def\BibTeX{{\rm B\kern-.05em{\sc i\kern-.025em b}\kern-.08em
    T\kern-.1667em\lower.7ex\hbox{E}\kern-.125emX}}
\begin{document}

\title{Mask-guided BERT for Few Shot Text Classification}

\author{
	\IEEEauthorblockN{1\textsuperscript{st}  Wenxiong Liao}
	\IEEEauthorblockA{\textit{South China University of Technology} \\
		cswxliao@mail.scut.edu.cn}
	\and
	
	\IEEEauthorblockN{2\textsuperscript{nd} Zhengliang Liu}
	\IEEEauthorblockA{\textit{University of Georgia} \\
		zl18864@uga.edu}
	\and
	
	\IEEEauthorblockN{3\textsuperscript{rd} Haixing Dai}
	\IEEEauthorblockA{\textit{University of Georgia} \\
		Haixing.Dai@uga.edu}
	\and
	
	\IEEEauthorblockN{4\textsuperscript{th} Zihao Wu}
	\IEEEauthorblockA{\textit{University of Georgia} \\
		zw63397@uga.edu}
	\and
	
	\IEEEauthorblockN{5\textsuperscript{th} Yiyang Zhang}
	\IEEEauthorblockA{\textit{South China University of Technology} \\
		zyyinyourarea@163.com}
	\and
	
	\IEEEauthorblockN{6\textsuperscript{th}Xiaoke Huang}
	\IEEEauthorblockA{\textit{South China University of Technology} \\
		csxkhuang@mail.scut.edu.cn}
	\and
	
	\IEEEauthorblockN{7\textsuperscript{th}Yuzhong Chen}
	\IEEEauthorblockA{\textit{University of Electronic Science and Technology} \\
		chenyuzhong211@gmail.com}
	\and

	\IEEEauthorblockN{8\textsuperscript{th}Xi Jiang}
	\IEEEauthorblockA{\textit{University of Electronic Science and Technology} \\
		xijiang@uestc.edu.cn}
	\and
	
	\IEEEauthorblockN{9\textsuperscript{th}Wei Liu}
	\IEEEauthorblockA{\textit{Mayo Clinic} \\
		liu.wei@mayo.edu}
	\and
	
	\IEEEauthorblockN{10\textsuperscript{th}Dajiang Zhu}
	\IEEEauthorblockA{\textit{University of Texas at Arlington} \\
		dajiang.zhu@uta.edu}
	\and
	
	\IEEEauthorblockN{11\textsuperscript{th} Tianming Liu}
	\IEEEauthorblockA{\textit{University of Georgia} \\
		tliu@uga.edu }
	\and
	
	\IEEEauthorblockN{12\textsuperscript{th} Sheng Li}
	\IEEEauthorblockA{\textit{University of Virginia} \\
		vga8uf@virginia.edu}
	\and
	
	\IEEEauthorblockN{13\textsuperscript{th} Xiang Li}
	\IEEEauthorblockA{\textit{Massachusetts General Hospital and Harvard Medical School} \\
		xli60@mgh.harvard.edu}
	\and

	\IEEEauthorblockN{14\textsuperscript{st} Hongmin Cai}
	\IEEEauthorblockA{\textit{South China University of Technology} \\
		hmcai@scut.edu.cn}
	\and

}

\maketitle

\begin{abstract}
	
	Transformer-based language models have achieved significant success in various domains. However, the data-intensive nature of the transformer architecture requires much labeled data, which is challenging in low-resource scenarios (i.e., few-shot learning (FSL)). The main challenge of FSL is the difficulty of training robust models on small amounts of samples, which frequently leads to overfitting. Here we present Mask-BERT, a simple and modular framework to help BERT-based architectures tackle FSL. The proposed approach fundamentally differs from existing FSL strategies such as prompt tuning and meta-learning. The core idea is to selectively apply masks on text inputs and filter out irrelevant information, which guides the model to focus on discriminative tokens that influence prediction results. In addition, to make the text representations from different categories more separable and the text representations from the same category more compact, we introduce a contrastive learning loss function. Experimental results on public-domain benchmark datasets demonstrate the effectiveness of Mask-BERT. 
\end{abstract}

\section{Introduction}
The transformer~\cite{vaswani2017attention} has become the standard architecture for various natural language processing (NLP) and computer vision tasks ~\cite{qiu2020pre,lin2022survey,kalyan2021ammus}. However, despite the dominating success and ubiquitous application of transformers, the majority of transformer-based models (e.g., BERT ~\cite{devlin2019bert} and Vision Transformer ~\cite{dosovitskiy2020vit}) need to be trained with substantial amounts of labeled data ~\cite{parnami2022learning, rezayiagribert, dosovitskiy2020vit}. In many cases, it is difficult or even impossible to supply sufficient data to such data-hungry pipelines ~\cite{mueller2022label, rezayiagribert}. For example, it is often infeasible to collect and annotate more than a few hundred samples in clinical practices ~\cite{ge2022few}. In response, research on few-shot learning (FSL) aims to develop models that could generalize from a few labeled instances and deliver acceptable performance in the downstream tasks ~\cite{yang2021survey, beltagy2022zero}. 

To effectively address the FSL challenge in NLP, many approaches have been proposed in the literature. Generally speaking, existing methods target either model design ~\cite{sun2019hierarchical, yin2020meta, wang2021transprompt}, data augmentation ~\cite{wei2019eda, kumar2019closer} or training strategies ~\cite{wei2021few}. In many cases, it is also possible to combine data augmentation methods with dedicated few-shot models ~\cite{wei2021few, wang2021entailment}. Some of the most prominent ~\cite{yin2020meta, yang2021survey, ge2022few} approaches that have gained traction include meta-learning ~\cite{yin2020meta, lee2022meta} and prompting ~\cite{brown2020language, lester2021power, han2022ptr, wang2022towards}. 


Despite their recent success, prompting and meta-learning have a few major limitations. For example, the manual design of prompts is cumbersome and prone to suboptimal designs ~\cite{gao2021making}. Some state-of-the-art prompt tuning methods (e.g., ~\cite{lester2021power, li2021prefix, hambardzumyan2021warp}) also require prompt engineering (e.g., through hyper-parameters) ~\cite{liu2021pre}. The prompts generated by prompt-tuning methods can also be unintelligible to humans, which impedes further manipulations ~\cite{liu2021pre}. Meta-learning, on the other hand, suffers from limitations such as the difficulty of finding appropriate meta-parameters ~\cite{lee2022meta}, training instability ~\cite{antoniou2018train, finn2017model, yao2021model}, and sensitivity to hyper-parameters and random seeds ~\cite{antoniou2018train, finn2017model}. More importantly, the design and implementation of both prompting and meta-learning models are quite complex, requiring extensive expertise in machine learning and/or the problem domain. Such complexity often hinders the practical applicability of these models. 

Inspired by research in neuroscience ~\cite{eugster2014predicting, liang2015effects, fitzsimmons2020impact, ye2021understanding} and recent advancement on \textit{BERTology} (i.e., the study of BERT and its variants' inherent properties and behaviors) ~\cite{tenney2019bert, clark2019does, rogers2020primer}, we propose the Mask-BERT framework to enhance few-shot learning capabilities of BERT-based models. The primary novelties and contributions of our work are as follows:

\begin{itemize}
	\item A simple and modular framework (Mask-BERT) is proposed for enhancing the few-shot learning performance of BERT-based language models, which is completely different from existing prompting and meta-learning methods and achieves superior performance in multiple tasks. 
	\item A novel masking strategy: Mask-BERT learns and applies a mask on the input text to filter out irrelevant information and directs the model's attention to more discriminative tokens to facilitate FSL. 
	\item We utilize anchor samples from the source domain data and adopt a contrastive learning scheme for FSL within the Mask-BERT framework. Experimental results indicate their effectiveness in improving model generalizability in FSL.
\end{itemize}

\section{Related Work}

\subsection{Masking model inputs}
Masking inputs is an inspiration from the computer vision community. In the seminal work of ~\cite{he2022masked}, the authors propose a simple framework where an autoencoder is fed with partially masked images, and the accompanying decoder is tasked with reconstructing the original images. Inspired by the work of ~\cite{chen2022mask}, we envision that applying mask operation on model input can effectively screen out task-irrelevant information from the input text and guide the model to focus on task-relevant and discriminative tokens.

In addition, prior research indicates that humans focus on \textit{key information} while reading ~\cite{chi2005scenthighlights, xu2013text, ye2021understanding}. There is also evidence indicating that highly relevant terms elicit stronger responses in certain brain areas ~\cite{eugster2014predicting, ye2021understanding}. In other words, the brain pays uneven attention to different sentence segments. We incorporate this biologically-inspired design into BERT, a descendant from the line of research on attention mechanism ~\cite{vaswani2017attention}. 

\subsection{Few-Shot Learning in NLP}
FSL methods of transformer-based pre-trained language models, especially those targeted at text classification, can be roughly divided into three categories: prompt-based methods, meta-learning methods, and fine-tunning based methods. By definition, the proposed Mask-BERT model belongs to the category of fine-tunning based methods.

The prompt-based methods include token-level prompt-learning and sentence-level prompt learning. Token-level prompt-learning is based on the pre-training task of the masked language model, while sentence-level prompt-learning is based on the pre-training task of next sentence prediction. PET ~\cite{schick2021exploiting,schick2021s} is commonly used in token-level prompt-learning methods, and ~\cite{sun2022nsp} propose a state-of-the-art sentence-level prompt-learning method named NSP-BERT for various NLP few-shot learning tasks.

The meta-learning methods include optimization-based meta-learning ~\cite{lei2022adaptive} and metric-based meta-learning. The siamese neural networks ~\cite{koch2015siamese}, prototypical network ~\cite{snell2017prototypical}, and induction network ~\cite{geng2019induction} are commonly used in metric-based meta-learning approaches. ~\cite{mueller2022few} propose a concise and efficient model based on siamese networks and label tuning for few-shot text classification. 

The fine-tunning based methods replace the original output layer of pre-trained language models with a new task-specific layer and fine-tunes the model using few-shot learning samples. Here we focus on the fine-tuning of BERT ~\cite{devlin2019bert}. Since BERT is pre-trained on open-domain datasets, which have different data distributions from the task-specific target domains, further pre-training of BERT with target domain data is also helpful for domain adaption ~\cite{sun2019fine}. ~\cite{zhang2020revisiting} shows that re-initializing top pre-trained layers help improve the fine-tuning performance. However, it is difficult to train all BERT parameters in the context of FSL where samples are extremely limited. In contrast, our proposed Mask-BERT model only adds the mask operation to BERT, and will inherit parameters from pre-trained models without any further costly pre-training.

\subsection{Contrastive learning}

Contrastive learning ~\cite{hadsell2006dimensionality} is a similarity-based learning strategy that has been widely used in visual representation ~\cite{he2020momentum}, 
graph representation ~\cite{qiu2020gcc}  and NLP ~\cite{gao2021simcse} tasks. ~\cite{li2021knn} introduce a supervised momentum contrastive learning framework to learn the clustered representations of text samples and utilize the K-nearest neighbors classifier to fine-tune pre-trained models ~\cite{li2021knn}. \cite{zhang2021few} propose a few-shot intent detection framework based on fine-tuning and contrastive learning, which performs self-supervised contrastive pre-training on abundant intent datasets at first and then conducts few-shot intent detection with supervised contrastive learning. \cite{das2022container} propose a contrastive learning framework via Gaussian embedding and optimizes the inter-token distribution distance for few-shot named entity recognition. Inspired by previous research, we adopt contrastive learning and exploit anchor samples to make representations of different text categories more separable and representations of the same category more compact.

\begin{figure*}[t]
	\centering
	\includegraphics[width=1.0\textwidth]{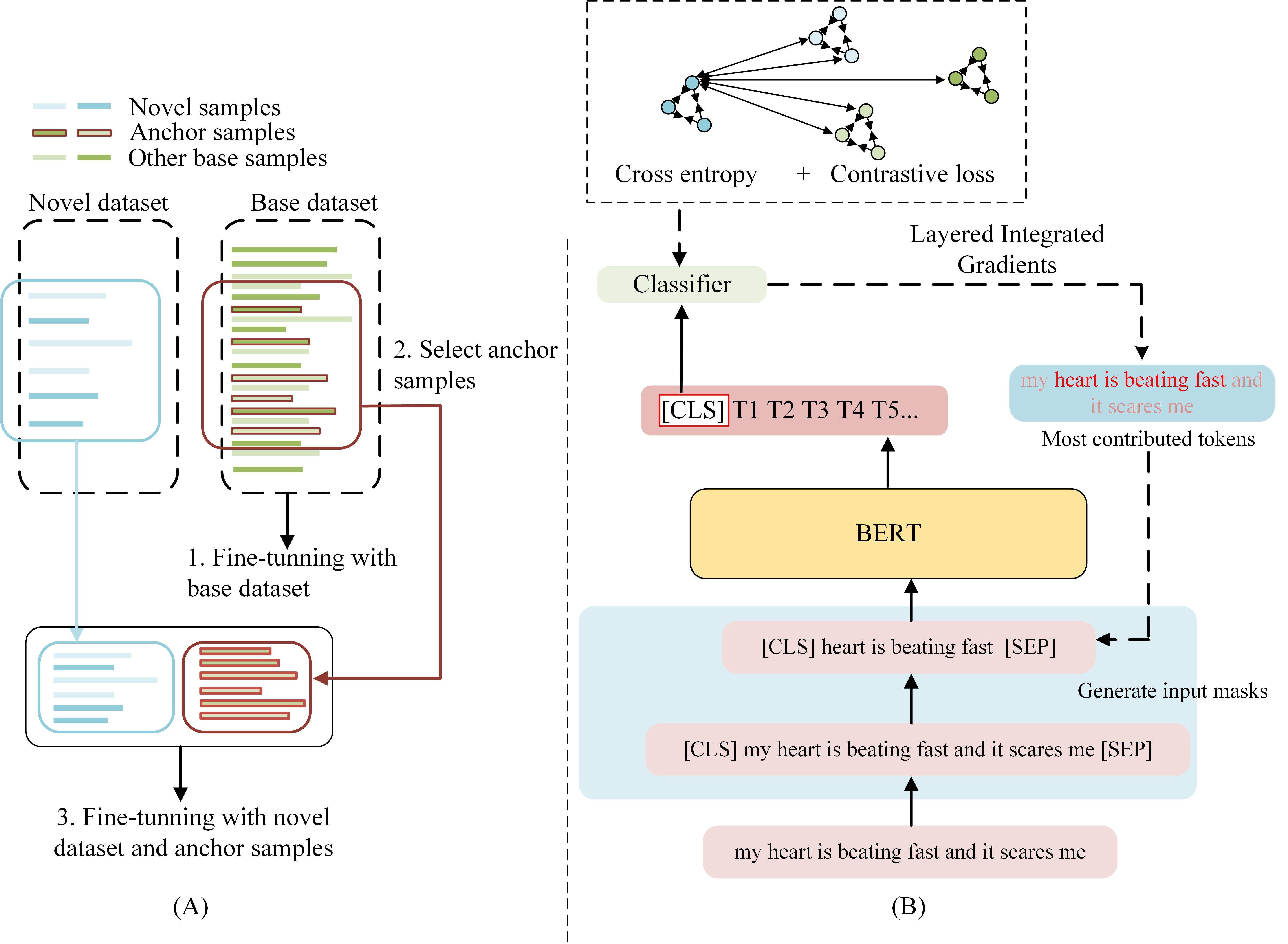} 
	\caption{(A). The overall framework of Mask-BERT. This framework includes three steps: 1) Fine-tune the parameters of BERT on the base dataset. 2) Select anchor samples from the base dataset. 3) Take few-shot novel samples and anchor samples to fine-tune the model. (B). Generate input masks for anchor samples. The objective function is based on the cross entropy loss and the contrastive loss.}
	\label{fig1}
\end{figure*}

\section{Methodology}
In this section, we first define our task and introduce the pre-training and fine-tuning paradigm of BERT. Then, we summarize the overall Mask-BERT pipeline.  Finally, we describe the Mask-BERT architecture in detail, including anchor samples selection, input mask generation, and the objective function.


\subsection{Problem definition}
Given a base dataset $D_b = \{(x_i,y_i)\}_{i=1}^{N_{b}}$ where $y_i \in Y_b$ and a novel dataset  $D_n = \{(x_j,y_j)\}_{j=1}^{N_{n}}$ where $y_j \in Y_n$.  The $D_b$ and $D_n$ have a disjoint label space, i.e. $Y_b \cap Y_n = \emptyset$. In our few-shot learning scenario, the base dataset contains a large number of labeled samples, while the novel dataset contains only a few labeled samples. Our goal is to use the model trained on the base dataset to achieve satisfactory generalization performance on the novel dataset through FSL. If the label space of novel dataset $D_n$ contains $N$ categories and each category contains $K$ labeled samples, this task is defined as a $N$-way-$K$-shot task. 

\subsection{BERT for sentence classification}
The BERT model \cite{devlin2019bert} consists of transformer encoder blocks. In the context of BERT, a sentence is defined as:
\begin{align}\label{eq1}
x_i = [w_{CLS}, w_1, \cdots, w_n]
\end{align}
where $w_{CLS}$ is regarded as the class special token. The sentence maps to a sequence of continuous representation:
\begin{align}\label{eq2}
z^0 = [z_{CLS}^0,z_1^0,...,z_n^0]
\end{align}

Each layer of BERT consists of a multi-head self-attention (MHSA) block and a multi-layer perceptron (MLP) block. LayerNorm (LN) and residual connections are applied before and after each block, respectively. The output features $z^l$ of the $l$-th layer can be written as:
\begin{align}\label{eq3}
z^{l'} = MHSA(LN(z^{l-1})) + z^{l-1}, l = 1,...,L
\end{align}
\begin{align}\label{eq4}
z^{l} = MLP(LN(z^{l'})) + z^{l'}, l = 1,...,L
\end{align}

For text classification, the features of the special token at the top layer (i.e. $z_{CLS}^L$) are usually fed into the task-specific MLP header for final prediction. 

BERT follows the pre-training and fine-tuning paradigm. The model is pre-trainied on a large-scale dataset and then fine-tuned on relatively small-scale (such as 5000 samples) labeled data associated with specific tasks. However, sometimes it is difficult to obtain sufficient labeled data for fine-tuning and achieve satisfactory performance.

\begin{algorithm}[tb]
	\caption{The Mask-BERT framework.}
	\label{algorithm_1}
	\textbf{Input}:  Base dataset ${D}_b$ and novel dataset ${D}_n$ \\
	\textbf{Initialize}: Initialized pre-trained BERT  $model$ \\
	\textbf{Parameters}: Fine-tuning epochs of base dataset $epoch_b$, fine-tuning epochs of FSL $epoch_f$, mask ratio $ratio$ \\
	\textbf{Output}: FSL $model$
	\begin{algorithmic}
		\FOR{epoch {\bfseries in} $epoch_b$}
		\STATE train($model$, ($D_b$, mask=None))
		\ENDFOR
		\STATE $D_b^{sub}$ = get\_anchors(${D}_b$)
		\FOR{epoch {\bfseries in} $epoch_f$}
		\STATE Generate $mask_b^{sub}$ as Eq~\ref{eq5}
		\STATE $D_h$ = {($D_n$,None)$\cup${($D_b^{sub}$,$mask_b^{sub}$))}}
		\STATE train($model$, $D_h$)
		\ENDFOR
	\end{algorithmic}
\end{algorithm}

\subsection{The Mask-BERT framework}

We propose Mask-BERT for few-shot text classification. One of the core challenges of FSL is to effectively adapt prior knowledge learned from the source domain (base dataset) to the target domain (novel dataset). We design Mask-BERT to filter out task-irrelevant inputs and guide the model to focuson task-relevant and discriminative tokens.

The framework of Mask-BERT is shown in Figure \ref{fig1}, and the training steps of Mask-BERT are shown in Algorithm \ref{algorithm_1}. First, we load the pre-trained BERT model and fine-tune it on the base dataset ${D}_b$. Next, we select the sample subset $D_b^{sub}$ from ${D}_b$ as anchor samples, and calculate the corresponding mask $mask_b^{sub}$. Finally, we fine-tune the $model$ on the ${D}_n$ and $D_b^{sub}$ (with mask) together.

It should be noted that Mask-BERT only performs mask operation on the neighborhood samples $D_b^{sub}$ but not on the novel dataset for two reasons. First, we want to make full use of the few-shot sample information in the novel dataset. Second, it is difficult to identify the important features within only a few samples in the novel dataset, which may cause noise pollution. 

\subsection{Selection of anchor samples}

The sample size of the novel dataset is too small for typical language models to learn stable and robust representations, which easily leads to overfitting. Since anchor samples are selected from the larger base dataset, the representations of anchor samples are stable and can guide the modeling process of the novel dataset. We select anchor samples from the  base dataset based on the following two rules: 1). Anchor samples should be close to the class center to prevent the introduction of noisy samples. 2). In order to guide the classification of novel data and avoid increasing the difficulty of FSL on the novel dataset, anchor samples should be far away from novel data.

Specifically, we first use the BERT model (fine-tuned on the base dataset) as a feature extractor to extract text features of all base samples and the few-shot novel samples. Then we locate the class centers of each category in the base dataset and compute the distances $d_b$ between each base sample and their corresponding class centers. At the same time, we calculate the distances $d_n$ between every base sample and every few-shot novel sample. Finally, $K_b$ samples with the lowest $d_b - d_n$ values are selected from each category of the base dataset. We use the Euclidean distance as the distance metric. To balance the distribution of sample categories, we set $K_b = K$ .

\subsection{Generation of input masks}

In order to take full advantage of prior knowledge and reduce the bias between the source and target domains, we generate a token mask to select text fragments related to the novel dataset. We use Integrated Gradients ~\cite{sundararajan2017axiomatic}, a model-agnostic Explainable-AI method that computes attribution scores with regard to the input features. It helps explain prediction results in terms of the inputs, and provides insights into feature importance and contributions to predictions. 

In the context of text classification, we compute the attribution scores for each token in the input sequence. To ensure semantic coherence, we keep the contiguous text segments that contribute the most to the classification task, rather than discrete tokens. For ${\forall}(x_i) \in D_b^{sub}$, this operation can be expressed as: 
\begin{align}\label{eq5}
mask_i = mask\_generator(x_i,model,ratio) 
\end{align}
where the hyperparameter $ratio$ is the ratio of the number of tokens of the selected continuous text segment to the number of tokens of the entire sentence.

The mask operation can take better advantage of prior knowledge from the source domain and reduce the deviation between the source and target domains. At the same time, we dynamically mask low-contribution tokens in each iteration. Different masks can carry out feature perturbation of samples on the premise of maintaining the same sentence semantics, which increases the robustness of the model.

\subsection{Objective function}

We feed the features of the special token at the top layer (i.e. $z_{CLS}^L$ ) into a fully connected layer for final prediction and take the cross-entropy over the samples:
\begin{align}\label{eq6}
\hat{y} = W^T z_{CLS}^L + b
\end{align}
\begin{align}\label{eq7}
L_{cross} = - \sum_{d\in D_n \cup D_b^{sub} } \sum_{c=1}^C y_{dc} \ln \hat{y}_{dc}
\end{align}
where $C$ is the dimension of the output features, which is equal to the union of the label spaces of base dataset and novel dataset.

In order to make full use of the prior knowledge from the base dataset and take full advantage of anchor samples guidance, we apply a contrastive loss to increase the distances between feature representations of samples belonging to different categories and shorten the distances between feature representations of samples within the same category. The computation of this contrastive loss between pairs of samples in the same batch is defined as follows:
\begin{align}\label{eq8}
L_{ctra} =  - \log \frac{{\sum {e^{cos({z_i},{z_{i'}})}} }}{{\sum {e^{cos({z_i},{z_{i'}})}}  + \sum {e^{cos({z_i},z_j)}} }}\
\end{align}
where $z_i$ and $z_i^{'}$ are the  special token features $z_{CLS}^L$ of samples within the same category,  $x_i$ and $x_j$ are the $z_{CLS}^L$ of samples belonging to different categories. $cos$ represents cosine similarity.

The objective function of the whole framework is defined as:
\begin{align}\label{eq9}
L_{total} = L_{cross} + L_{ctra}.
\end{align}

\section{Experiments}

\begin{table}[t]
	\caption{Summary of datasets. }
	\label{table1}
	\resizebox{1.0\columnwidth}{!}{
		\begin{tabular}{llll}
			\hline
			Dataset      & Avg.length & \ $\lvert Y_b \rvert$/$ \lvert Y_n \rvert$ & domain         \\ \hline
			AG News      & 39         & 2/2 & open-domain    \\
			Dbpedia14    & 50         & 8/6 & open-domain    \\
			Snippets     & 18         & 4/4 & open-domain    \\
			Symptoms     & 11         & 4/3 & medical-domain \\
			PubMed20k & 26         & 3/2 & medical-domain \\
			NICTA-PIBOSO    & 24         & 3/2 & medical-domain \\ \hline
	\end{tabular}}
\end{table}

\begin{table*}[h]
	\centering
	\caption{Accuracy (Acc) of text classification on benchmark datasets with Mask-BERT and baselines }
	\label{table2}
	\resizebox{1.0\textwidth}{!}{\begin{tabular}{llllllll}
			\hline
			K-shot                  & models       & AG news              & Dbpedia14            & Snippets             & Symptoms             & PubMed20k            & NICTA-PIBOSO         \\ \hline
			\multirow{8}{*}{5-shot} & BERT~\cite{devlin2019bert}         & 0.753±0.095          & 0.948±0.029          & 0.852±0.029          & 0.782±0.04           & 0.845±0.027          & 0.696±0.054          \\
			& FPT-BERRT~\cite{sun2019fine}    & 0.779±0.035          & 0.963±0.026          & 0.888±0.028          & 0.800±0.112          & 0.864±0.030          & 0.727±0.042          \\
			& Re-init-BERT~\cite{zhang2020revisiting} & 0.762±0.024          & 0.941±0.029          & 0.799±0.061          & 0.830±0.047          & 0.862±0.019          & 0.702±0.063          \\
			& CPFT~\cite{zhang2021few}         & 0.768±0.047          & 0.973±0.013          & 0.853±0.029          & 0.876±0.044          & 0.848±0.028          & 0.723±0.068          \\
			& CNN-BERT~\cite{harly2022cnn}     & 0.698±0.057          & 0.960±0.020          & 0.864±0.031          & 0.818±0.085          & 0.879±0.028          & 0.718±0.079          \\
			& SN-FT~\cite{mueller2022few}        & 0.768±0.055          & 0.982±0.003          & 0.867±0.024          & 0.858±0.059          & 0.782±0.031          & 0.736±0.041          \\
			& NSP-BERT~\cite{sun2022nsp}     & 0.820±0.016          & 0.985±0.002          & 0.885±0.016          & 0.870±0.043          & 0.701±0.096          & 0.732±0.047          \\
			& Mask-BERT    & \textbf{0.821±0.040} & \textbf{0.986±0.004} & \textbf{0.892±0.016} & \textbf{0.888±0.027} & \textbf{0.894±0.012} & \textbf{0.755±0.043} \\ \hline
			\multirow{8}{*}{8-shot} & BERT~\cite{devlin2019bert}         & 0.790±0.063          & 0.979±0.004          & 0.888±0.021          & 0.855±0.066          & 0.875±0.024          & 0.714±0.038          \\
			& FPT-BERRT~\cite{sun2019fine}    & 0.801±0.035          & 0.978±0.012          & 0.903±0.023          & 0.848±0.094          & 0.870±0.014          & 0.726±0.048          \\
			& Re-init-BERT~\cite{zhang2020revisiting} & 0.789±0.026          & 0.964±0.011          & 0.863±0.028          & 0.882±0.046          & 0.880±0.008          & 0.738±0.034          \\
			& CPFT~\cite{zhang2021few}         & 0.816±0.021          & 0.978±0.002          & 0.888±0.017          & 0.936±0.029          & 0.871±0.022          & 0.776±0.046          \\
			& CNN-BERT~\cite{harly2022cnn}     & 0.731±0.042          & 0.980±0.003          & 0.887±0.023          & 0.876±0.063          & 0.894±0.011          & 0.732±0.058          \\
			& SN-FT~\cite{mueller2022few}        & 0.817±0.024          & 0.986±0.002          & 0.881±0.020          & 0.876±0.081          & 0.796±0.044          & 0.757±0.033          \\
			& NSP-BERT~\cite{sun2022nsp}     & 0.833±0.009          & 0.986±0.003          & 0.898±0.018          & 0.897±0.031          & 0.761±0.051          & 0.749±0.040          \\
			& Mask-BERT    & \textbf{0.835±0.015} & \textbf{0.988±0.001} & \textbf{0.909±0.010} & \textbf{0.948±0.056} & \textbf{0.901±0.008} & \textbf{0.781±0.033} \\ \hline
	\end{tabular}}
\end{table*}

\begin{figure*}[h]
	\centering
	\includegraphics[width=1.0\textwidth]{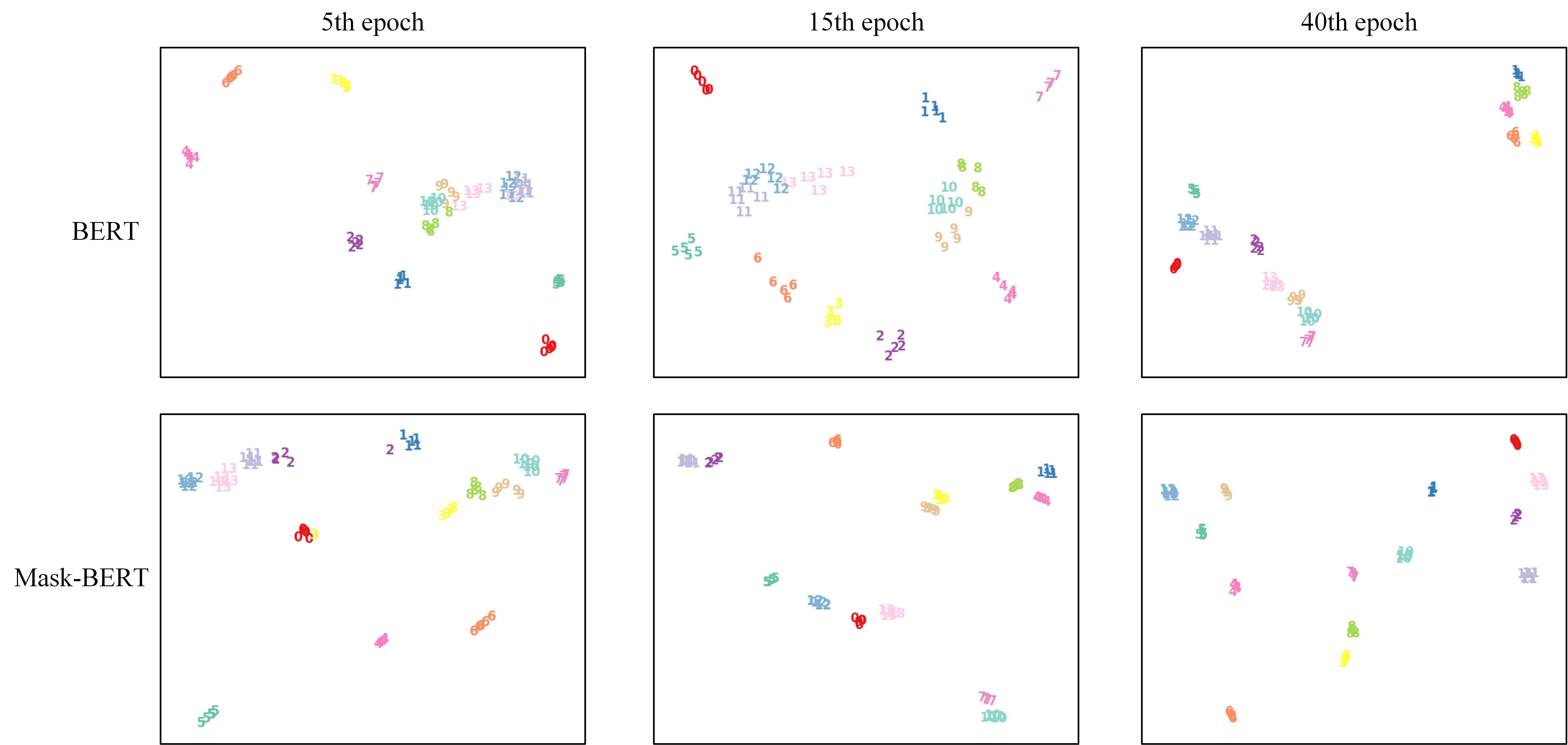} 
	\caption{Visualization of training iteration on Dbpedia14. }
	\label{fig2}
\end{figure*}

\begin{figure*}[h]
	\centering
	\includegraphics[width=1.0\textwidth]{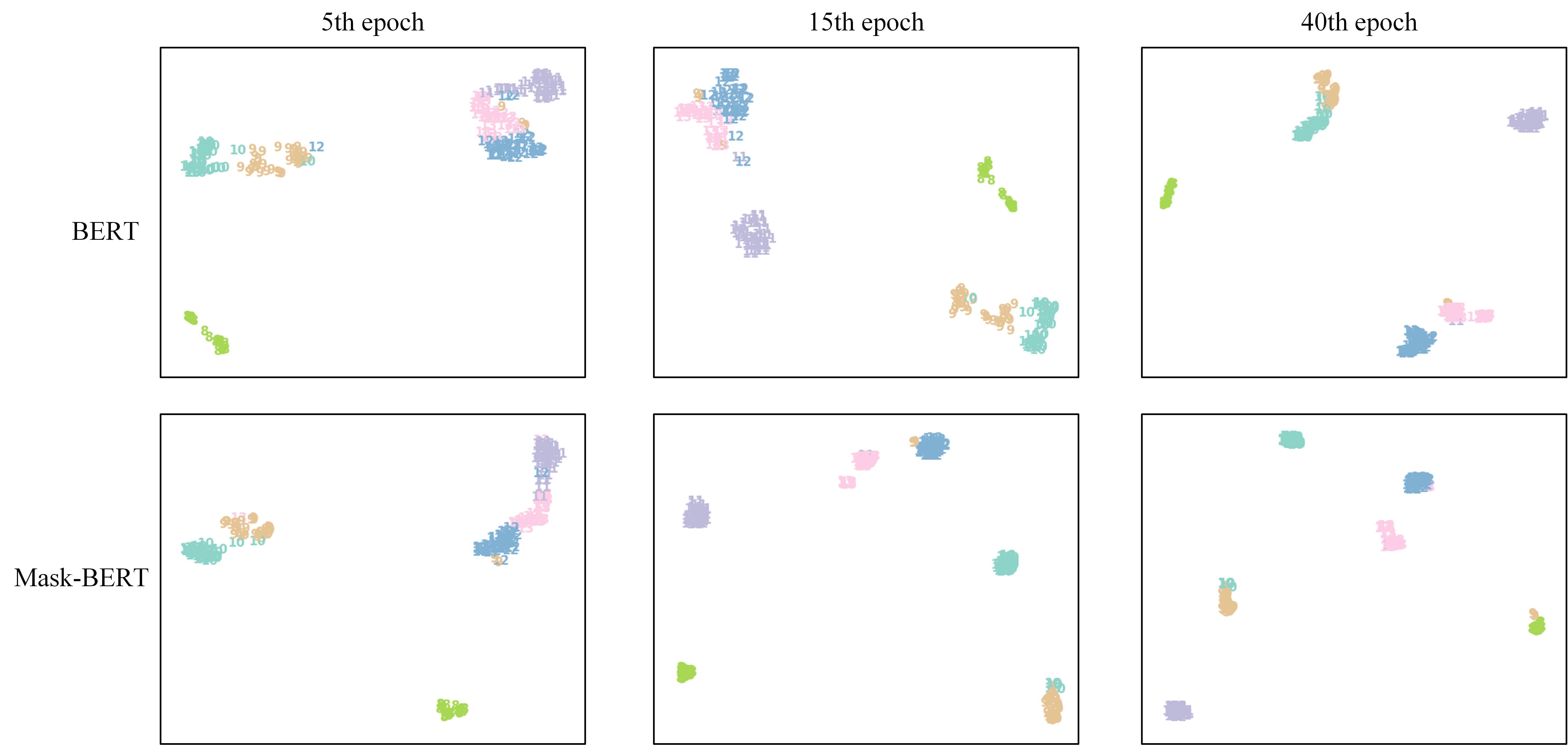} 
	\caption{Visualization of test set on Dbpedia14}
	\label{fig3}
\end{figure*}

In this section, we evaluate Mask-BERT on the text classification task and conduct experiments on publicly available benchmarks. We compare Mask-BERT against both baseline methods and state-of-the-art FSL models in all three families of few-shot learning methods (prompt-based FSL, meta-learning, and fine-tuning based methods). In addition, we conduct ablation studies on Mask-BERT to investigate the effectiveness of various model components. 


\subsection{Datasets}

Our experiment involves 6 public datasets, including 3 open-domain datasets and 3 medical-domain datasets. The average length of samples in each dataset and the number of categories in the base/novel datasets are summarized in Table \ref{table1}. 

\begin{itemize}
	
	\item \textbf{AG news}. This is a text classification dataset that contains 4 news categories
	~\cite{zhang2015character}.
	
	\item \textbf{DBpedia14}. This is a text classification dataset with 14 categories
	~\cite{zhang2015character}.
	
	\item \textbf{Snippets}. It is a dataset of web search snippets retrieved from Google Search
	~\cite{phan2008learning}
	
	\item \textbf{Symptoms}. This dataset is published on Kaggle\footnote{https://www.kaggle.com/datasets/paultimothymooney/medical-speech-transcription-and-intent}. It contains the audio data of common medical symptom descriptions over 8 hours. We use the text transcripts corresponding to the audio data and perform sample de-duplication. The dataset, after preprocessing, includes 231 samples of 7 symptom categories.
	
	\item \textbf{PubMed20k}. It is a dataset based on PubMed for sequence classification. 
	~\cite{dernoncourt2017pubmed}
	
	\item \textbf{NICTA-PIBOSO}. This dataset is based on the ALTA 2012 Shared Task. It is a benchmark for classifying sentences in biomedical abstracts into pre-defined categories.
	~\cite{amini2012overview}.
	

\end{itemize}

\subsection{Experimental Setup}
We compare Mask-BERT with recent NLP models, which are briefly summarized here:
\begin{itemize}
	\item \textbf{BERT} ~\cite{devlin2019bert} is the standard fine-tuning method of BERT.
	
	\item \textbf{FPT-BERT} (further pre-train BERT) ~\cite{sun2019fine}. FPT-BERT is further pre-trained  through masked language modeling.
	
	\item \textbf{Re-init-BERT}  ~\cite{zhang2020revisiting}. Re-init-BERT re-initializes the top layers of the BERT model.
	
	\item \textbf{CPFT} ~\cite{zhang2021few}. CPFT is a contrastive framework including contrastive pre-trained and contrastive fine-tuning.
	
	\item \textbf{CNN-BERT} ~\cite{harly2022cnn} applies CNN to classify text based on features outputted from the BERT model.
	
	\item \textbf{SN-FT} ~\cite{mueller2022few}. SN-FT is a metric-based meta-learning method with siamese networks for few-shot text classification.
	
	\item \textbf{NSP-BERT} ~\cite{sun2022nsp}. NSP-BERT is a state-of-the-art sentence-level prompt-learning method.
	
\end{itemize}

In Mask-BERT, we fine-tune BERT on the base dataset in all the tasks with  batch\_size = 64, except for the Symptoms dataset, where we use batch\_size = 32.
$epoch_b$ is set as 8 for all tasks. We randomly select $K$ ($K = 5, 8$) samples in each novel category from the training set as the novel dataset, and fine-tune the model on these few-shot samples (along with the anchor samples $D_b^{sub}$ from the base dataset). The $epoch_f$ is set as 50 for PubMed20k and Dbpedia14, and for other tasks it is set as 150. We use AdamW ~\cite{loshchilov2018decoupled} with a learning rate of 2e-5 and SignSGD ~\cite{bernstein2018signsgd} with a learning rate 4e-5 for fine-tuning on the base dataset and novel dataset, respectively. We apply grid search to determine the value of $ratio$ from 0.05 to 0.85 with an interval of 0.1. To ensure the objectivity of the experiment, we initialize all methods with the pre-trained bert-base-cased\footnote{https://huggingface.co/bert-base-cased} model. BERT, Re-init-BERT, CNN-BERT are fine-tuned based on the base dataset, similar to Mask-BERT. For FPT-BERT, we follow the original implementation and continue its pre-training on the base dataset. We use 10 different random seeds to select $K$ samples as the novel dataset and report the average accuracy of the test set with only novel classes.


\subsection{Experimental Results}
The main results of all the performance comparisons are listed in Table \ref{table2}. Overall,  Mask-BERT and NSP-BERT have similar superior performance on open-domain datasets. However, Mask-BERT achieves the best performance on medical-domain datasets. Compared to other fine-tuning methods, Mask-BERT performs well with just a few novel samples and improves the performance of the pre-trained language model in the few-shot learning scenario. In addition, it is found that when the mask $ratio$ is from 0.05 to 0.85, the Mask-BERT performance is quite stable regarding the mask $ratio$ for all the datasets. Quantitatively, the average standard deviation of the accuracy of Mask-BERT among different mask $ratio$ is 0.006 across all the datasets. 

\subsection{Result Analysis}

\begin{table}[t]
	\centering
	\caption{Ablation study of 5-shot task on AG news and NICTA-PIBOSO dataset }
	\label{table3}
	\resizebox{1.0\columnwidth}{!}{
		\begin{tabular}{lll}
			\hline
			methods                             & AG News              & NICTA-PIBOSO         \\ \hline
			BERT                                & 0.753±0.095          & 0.696±0.054          \\
			+ constractive                      & 0.764±0.072          & 0.707±0.050          \\
			+ anchor                            & 0.798±0.058          & 0.730±0.044          \\
			+ constractive + random anchor        & 0.807±0.040          & 0.732±0.046          \\
			+ constractive + anchor               & 0.809±0.052          & 0.740±0.044          \\
			+ constractive + anchor + random mask & 0.811±0.035          & 0.729±0.038          \\
			+ constractive + anchor +  mask       & \textbf{0.821±0.040} & \textbf{0.755±0.043} \\ \hline
		\end{tabular}
	}
\end{table}

In this section, we will investigate the impact of each component of Mask-BERT through ablation experiments. We will also visualize the process of Mask-BERT during text classification.

\paragraph{Ablation study.} 
To analyze the impact of various components of Mask-BERT on model performance, we conducted ablation studies using the 5-shot tasks on the AG news and NICTA-PIBOSO dataset. The results are shown in Table \ref{table3}. ``random anchor'' means randomly selecting samples from the base dataset as anchor samples.  ``random mask'' means masking tokens randomly rather than according to the contribution of tokens. Table \ref{table3} shows that: 1) Adding contrastive loss in the fine-tuning stage can make samples of the same category more compact and samples of different categories more distant, thus improving the performance of the model; 2) Introducing anchor samples can make full use of the stable representation learned in the source domain, thus improving the learning of novel data; 3) Mask operation can guide BERT to focus on task-relevant and critical features in the input text.





\paragraph{Intermediate Result Visualization}We visualize the training iteration of 5-shot on the DBpedia14 dataset in Figure \ref{fig2} by mapping the samples to 2-D by UMAP~\cite{mcinnes2018umap} and color-code the results by the class labels. In DBpedia14, the label id of the base dataset is from 0 to 7, and the label id of the novel dataset is from 8 to 13. For comparison, we also visualize the intermediate results from BERT. The visualization shows that while both BERT and Mask-BERT can effectively separate the samples from different classes during training, Mask-BERT can span the classes into a more even distribution and avoid ''clusters'' of samples as in BERT, thanks to the guidance of stable anchor samples and the contrastive loss. Such a feature is vital for the effectiveness of FSL, as illustrated in Figure \ref{fig3} where we randomly sampled 200 samples from the test (novel) dataset and performed the visualization in a similar approach. It can be found that certain classes in the novel dataset are difficult to be separated by BERT as all samples were distributed in two big clusters (note the results from BERT in the 15th and 40th epoch). In contrast, Mask-BERT can solve this issue and improve the compactness of samples within the same class.

\section{Conclusions}
In this paper, we proposed a simple and modular framework named Mask-BERT to improve the few-shot learning capabilities of BERT models. We use anchor samples with stable representations to guide the learning on novel datasets. Furthermore, we mask some tokens with a lower contribution of anchor samples to guide the model to focus on task-related features and improve robustness. The contrastive loss function is employed to enhance the compactness of the same categories and the separation of the different categories. By testing the model on several public datasets, we show the superior performance of Mask-BERT over multiple other NLP methods. Results from the ablation study further validate the effectiveness of various components of Mask-BERT.

\bibliography{axriv_latex}
\bibliographystyle{IEEEtran}
\end{document}